\documentstyle[epsfig,titlepage,twoside,times,myapa]{article}

\renewcommand{\cite}[1]{\shortcite{#1}}
\newcommand{\namecite}[1]{\shortciteA{#1}}
\newcommand{\opencite}[1]{\shortciteNP{#1}}

\makeatletter
\def\section{\@startsection {section}{1}{\z@}{-3.5ex plus-1ex minus
    -.2ex}{2.3ex plus.2ex}{\reset@font\normalsize\bf}}
\def\subsection{\@startsection{subsection}{2}{\z@}{-3.25ex plus-1ex
    minus-.2ex}{1.5ex plus.2ex}{\reset@font\normalsize\it}}
\def\subsubsection{\@startsection{subsubsection}{3}{\z@}{-3.25ex plus
    -1ex minus-.2ex}{1.5ex plus.2ex}{\reset@font\normalsize\it}}
\def\myparagraph{\@startsection
     {paragraph}{4}{\z@}{3.25ex plus1ex minus.2ex}{-1em}{\reset@font
     \normalsize\it}}
\makeatother
\renewcommand{\paragraph}[1]{\myparagraph{#1.}}

\title{Prosody-Based Automatic Segmentation of Speech into Sentences and Topics}

\author{
      Elizabeth Shriberg \quad Andreas Stolcke \\
      Speech Technology and Research Laboratory \\
      SRI International, Menlo Park, CA, U.S.A. \\
      \{ees,stolcke\}@speech.sri.com
      \and
      Dilek Hakkani-T\"ur \quad G\"okhan T\"ur \\ 
      Department of Computer Engineering, Bilkent University \\
      Ankara, 06533, Turkey \\
      \{hakkani,tur\}@cs.bilkent.edu.tr
}

\begin{document}

\maketitle

\sloppy 

\newcommand{\WT}{W_{\mbox{\scriptsize t}}}
\newcommand{\PLM}{P_{\mbox{\scriptsize LM}}}
\newcommand{\PDT}{P_{\mbox{\scriptsize DT}}}
\newcommand{\PHMM}{P_{\mbox{\scriptsize HMM}}}
\newcommand{\argmax}{\mathop{\rm argmax}}

\begin{center} 
    \bf Abstract
\end{center}
A crucial step in processing speech audio data for information
extraction, topic detection, or browsing/playback is to segment the
input into sentence and topic units.  Speech segmentation is
challenging, since the cues typically present for segmenting text
(headers, paragraphs, punctuation) are absent in spoken language.  We
investigate the use of prosody (information gleaned from the timing
and melody of speech) for these tasks.  Using decision tree and hidden
Markov modeling techniques, we combine prosodic cues with word-based
approaches, and evaluate performance on two speech corpora, Broadcast
News and Switchboard.  Results show that the prosodic model alone
performs on par with, or better than, word-based statistical language
models---for both true and automatically recognized words in news
speech.  The prosodic model achieves comparable performance with
significantly less training data, and requires no hand-labeling of
prosodic events.  Across tasks and corpora, we obtain a significant
improvement over word-only models using a probabilistic combination of
prosodic and lexical information.  Inspection reveals that the
prosodic models capture language-independent boundary indicators
described in the literature. Finally, cue usage is task and corpus
dependent. For example, pause and pitch features are highly
informative for segmenting news speech, whereas pause, duration and
word-based cues dominate for natural conversation.

\newpage
\begin{center} 
    \bf Zusammenfassung
\end{center}
Ein wesentlicher Schritt in der Sprachverarbeitung zum Zweck der
Informations\-extra\-hierung, Themenklassifizierung oder Wiedergabe ist die
Segmentierung in themati\-sche und Satzeinheiten.  Sprachsegmentierung
ist schwierig, da die Hinweise, die daf\"ur gew\"ohn\-lich in Texten
vorzufinden sind (\"Uberschriften, Abs\"atze, Interpunktion), in
gespro\-chener Sprache fehlen.  Wir untersuchen die Benutzung von
Prosodie (Timing und Melodie der Sprache) zu diesem Zweck.
Mithilfe von Entscheidungsb\"aumen und Hidden-Markov-Modellen
kombinieren wir prosodische und wortbasierte Informationen, und
pr\"ufen unsere Verfahren anhand von zwei Sprachkorpora, Broadcast News
und Switchboard.  Sowohl bei korrekten, als auch bei automatisch
erkannten Worttranskriptionen von Broadcast News zeigen unsere
Ergebnisse, da{\ss} Prosodiemodelle alleine eine gleich\-gute oder
bessere Leistung als die wortbasieren statistischen Sprachmodelle
erbringen.  Dabei erzielt das Prosodiemodell eine vergleichbare
Leistung mit wesentlich weniger Trainingsdaten und bedarf keines
manuellen Transkribierens prosodischer Eigenschaften.  F\"ur beide
Segmentierungsarten und Korpora erzielen wir eine signifikante
Verbesserung gegen\"uber rein wortbasierten Modellen, indem wir
prosodische und lexikalische Informationsquellen probabilistisch
kombinieren.  Eine Untersuchung der Prosodiemodelle zeigt, da{\ss}
diese auf sprachunabh\"angige, in der Literatur beschriebene
Segmentierungsmerkmale ansprechen.  Die Auswahl der Merkmale h\"angt
wesentlich von Segmentierungstyp und Korpus ab.  Zum Beispiel sind
Pausen und F0-Merkmale vor allem f\"ur Nachrich\-ten\-sprache informativ,
w\"ahrend zeitdauer- und wortbasierte Merkmale in nat\"urlichen
Gespr\"achen dominieren.
\newpage

\begin{center} 
    \bf Resum\'e
\end{center}
Une \'etape cruciale dans le traitement de la parole pour l'extraction
d'information, la d\'etection du sujet de conversation et la
navigation est la segmentation du discours. Celle-ci est difficile car
les indices aidant \`a segmenter un texte (en-t\^etes, paragraphes,
ponctuation) n'apparaissent pas dans le language parl\'e.  Nous
\'etudions l'usage de la prosodie (l'information extraite du rythme et
de la m\'elodie de la parole) \`a cet effet. A l'aide d'arbres de
d\'ecision et de cha{\^\i}nes de Markov cach\'ees, nous combinons les
indices prosodiques avec le mod\`ele du langage. Nous evaluons notre
algorithme sur deux corpora, Broadcast News et Switchboard.
Nos r\'esultats indiquent que le mod\`ele prosodique est \'equivalent
ou sup\'erieur au mod\`ele du langage, et qu'il requiert moins de
donn\'ees d'entra{\^\i}nement. Il ne n\'ecessite pas d'annotations
manuelles de la prosodie. De plus, nous obtenons un gain significatif
en combinant de mani\`ere probabiliste l'information prosodique et
lexicale, et ce pour diff\'erents corpora et applications.  Une
inspection plus d\'etaill\'ee des r\'esultats r\'ev\`ele que les
mod\`eles prosodiques identifient les indicateurs de d\'ebut et de fin
de segments, tel que d\'ecrit dans la litterature.
Finalement, l'usage des indices prosodiques d\'epend de l'application
et du corpus. Par exemple, le ton s'av\`ere extr\`emement utile pour
la segmentation des bulletins t\'el\'evis\'es, alors que les
caracteristiques de dur\'ee et celles extraites du mod\`ele du
langage servent davantage pour la segmentation de conversations
naturelles.

\newpage

\twocolumn

\section{Introduction}
\subsection{Why process audio data?}

Extracting information from audio data allows examination of a much
wider range of data sources than does text alone. Many sources (e.g.,
interviews, conversations, news broadcasts) are available only in
audio form. Furthermore, audio data is often a much richer source than
text alone, especially if the data was originally meant to be {\em
  heard} rather than read (e.g., news broadcasts).

\subsection{Why automatic segmentation?}

Past automatic information extraction systems have depended mostly on
lexical information for segmentation
(\opencite{KubalaEtAl:98,Allan:98,Hearst:97,Kozima:93,Yamron:98},
among others).  A
problem for the text-based approach, when applied to speech input, is
the lack of typographic cues (such as headers, paragraphs, sentence
punctuation, and capitalization) in continuous speech.

A crucial step toward robust information extraction from speech is the
automatic determination of topic, sentence, and phrase boundaries.
Such locations are overt in text (via punctuation, capitalization,
formatting) but are absent or ``hidden'' in speech output.  Topic
boundaries are an important prerequisite for topic detection, topic
tracking, and summarization.  They are further helpful for
constraining other tasks such as coreference resolution (e.g., since
anaphoric references do not cross topic boundaries).  Finding sentence
boundaries is a necessary first step for topic segmentation.  It is
also necessary to break up long stretches of audio data prior to
parsing.  In addition, modeling of sentence boundaries can
benefit named entity extraction from automatic speech recognition
(ASR) output, for example by preventing proper nouns spanning a
sentence boundary from being grouped together.

\subsection{Why use prosody?} \label{sec:why-use-prosody}

When spoken language is converted via ASR to a simple stream of words,
the timing and pitch patterns are lost. Such patterns (and other
related aspects that are independent of the words) are known as speech
{\em prosody}. In all languages, prosody is used to convey structural,
semantic, and functional information.

Prosodic cues are known to be relevant to discourse structure across
languages (e.g., \opencite{Vaissiere:83}) and can therefore be expected to
play an important role in various information extraction tasks.
Analyses of read or spontaneous monologues in linguistics and related
fields have shown that information units, such as sentences and
paragraphs, are often demarcated prosodically. In English and related
languages, such prosodic indicators include pausing, changes in pitch
range and amplitude, global pitch declination, melody and boundary
tone distribution, and speaking rate variation.  For example, both
sentence boundaries and paragraph or topic boundaries are often marked
by some combination of a long pause, a preceding final low boundary
tone, and a pitch range reset, among other features
\cite{Lehiste:79,%
Lehiste:80,%
BrownEtAl:80,%
Bruce:82,%
Thorsen:85,%
Silverman:87,%
Grosz:92,%
Sluijter:94,%
Swerts:94,%
KoopmansDonzel:96,%
HirschbergNakatani:96,%
Nakajima:97,%
Swerts:97,%
SweOst:97}.

Furthermore, prosodic cues by their nature are relatively unaffected
by word identity, and should therefore improve the robustness of
lexical information extraction methods based on ASR output.  This may
be particularly important for spontaneous human-human conversation
since ASR word error rates remain much higher for these corpora than
for read, constrained, or computer-directed speech \cite{LVCSR:99}.

A related reason to use prosodic information is that certain prosodic
features can be computed even in the absence of availability of ASR,
for example, for a new language where one may not have a dictionary
available.  Here they could be applied for instance for audio browsing
and playback, or to cut waveforms prior to recognition to limit audio
segments to durations feasible for decoding.

Furthermore, unlike spectral features, some prosodic features (e.g.,
duration and intonation patterns) are largely invariant to changes in
channel characteristics (to the extent that they can be adequately
extracted from the signal). Thus, the research results are independent
of characteristics of the communication channel, implying that the
benefits of prosody are significant across multiple applications.

Finally, prosodic feature extraction can be achieved with minimal
additional computational load and no additional training data;
results can be integrated directly with existing conventional ASR
language and acoustic models.  Thus, performance gains can be
evaluated quickly and cheaply, without requiring additional
infrastructure.

\subsection{This study}

Past studies involving prosodic information have generally relied on
hand-coded cues (an exception is \opencite{HirschbergNakatani:96}).
We believe the present work
to be the first that combines fully automatic extraction of both
lexical and prosodic information for speech segmentation. 
Our general framework for combining lexical and prosodic cues for
tagging speech with various kinds of {\em hidden} structural
information is a further development of earlier work on detecting
sentence boundaries and disfluencies in spontaneous speech
\cite{%
ShribergEtAl:eurospeech97,%
SOBS:icslp98,%
DilekEtAl:euro99,%
StolckeEtAl:darpa99,%
TurEtAl:CL2000} and on detecting topic boundaries in Broadcast
News
\cite{%
DilekEtAl:euro99,%
StolckeEtAl:darpa99,%
TurEtAl:CL2000}.  In previous work we provided only a
high-level summary of the prosody modeling, focusing instead on
detailing the language modeling and model combination.

In this paper we describe the prosodic modeling in detail.  In
addition we include, for the first time, controlled comparisons for
speech data from two corpora differing greatly in style: Broadcast
News \cite{Graff:97} and Switchboard \cite{SWBD}.
The two corpora are
compared directly on the task of sentence segmentation, and the two
tasks (sentence and topic segmentation) are compared for the Broadcast
News data. Throughout, our paradigm holds the candidate features for
prosodic modeling constant across tasks and corpora. That is, we
created parallel prosodic databases for both corpora, and used the
same machine learning approach for prosodic modeling in all cases.  We
look at results for both true words, and words as hypothesized by a
speech recognizer.  Both conditions provide informative data points.
True words reflect the inherent additional value of prosodic
information above and beyond perfect word information. Using
recognized words allows comparison of degradation of the prosodic
model to that of a language model, and also allows us to assess
realistic performance of the prosodic model when word boundary
information must be extracted based on incorrect hypotheses rather
than forced alignments.

Section~\ref{sec:method} describes the methodology, including the
prosodic modeling using decision trees, the language modeling, the
model combination approaches, and the data sets.  The prosodic
modeling section is particularly detailed, outlining the motivation
for each of the prosodic features and specifying their extraction,
computation, and normalization.  Section~\ref{sec:results-discussion}
discusses results for each of our three tasks: sentence segmentation
for Broadcast News, sentence segmentation for Switchboard, and topic
segmentation for Broadcast News.  For each task, we examine results
from combining the prosodic information with language model
information, using both transcribed and recognized words. We focus on
overall performance, and on analysis of which prosodic features prove
most useful for each task. The section closes with a general
discussion of cross-task comparisons, and issues for further work.
Finally, in Section~\ref{sec:summary-conclusion} we summarize main
insights gained from the study, concluding with points on the general
relevance of prosody for automatic segmentation of spoken audio.

\section{Method} \label{sec:method}
\subsection{Prosodic modeling}
\subsubsection{Feature extraction regions}

\begin{figure*}
\begin{center}
\strut\psfig{figure=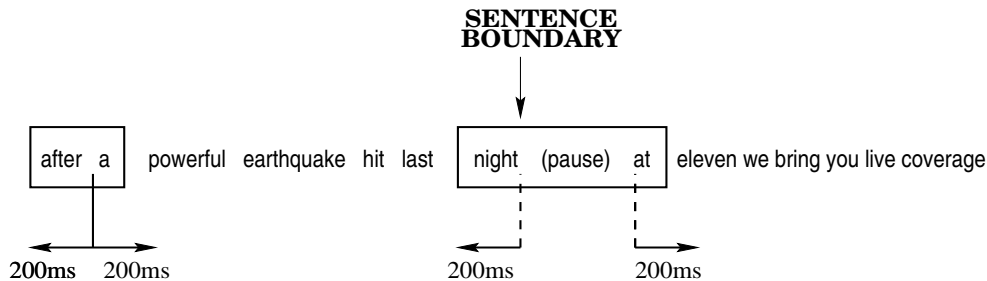,width=5.7in}
\end{center}
\caption{Feature extraction regions for each inter-word boundary}
\label{fig:extraction}
\end{figure*}

In all cases we used only very local features, for practical reasons
(simplicity, computational constraints, extension to other tasks),
although in principle one could look at longer regions.  As shown in
Fig.~\ref{fig:extraction}, for each inter-word boundary, we looked
at prosodic features of the word immediately preceding and following
the boundary, or alternatively within a window of 20 frames (200 ms, a
value empirically optimized for this work) before and after the
boundary. In boundaries containing a pause, the window extended
backward from the pause start, and forward from the pause end.  (Of
course, it is conceivable that a more effective region could be based
on information about syllables and stress patterns, for example,
extending backward and forward until a stressed syllable is reached.
However, the recognizer used did not model stress, so we preferred the
simpler, word-based criterion used here.)

We extracted prosodic features reflecting pause durations, phone
durations, pitch information, and voice quality information.  Pause
features were extracted at the inter-word boundaries.  Duration, F0,
and voice quality features were extracted mainly from the word or
window {\em preceding} the boundary (which was found to carry more
prosodic information for these tasks than the speech {\em following}
the boundary; \opencite{ShribergEtAl:eurospeech97}). We also included
pitch-related features reflecting the difference in pitch range {\em
  across} the boundary.

In addition, we included nonprosodic features that are inherently
related to the prosodic features, for example, features that make a
prosodic feature undefined (such as speaker turn boundaries) or that
would show up if we had not normalized appropriately (such as gender,
in the case of F0).  This allowed us both to better understand feature
interactions, and to check for appropriateness of normalization
schemes.

We chose not to use amplitude- or energy-based features, since
previous work showed these features to be both less reliable than and
largely redundant with duration and pitch features.  A main reason for
the lack of robustness of the energy cues was the high degree of
channel variability in both corpora examined, even after application
of various normalization techniques based on the signal-to-noise ratio
distribution characteristics of, for example, a conversation side (the
speech recorded from one speaker in the two-party conversation) in
Switchboard.  Exploratory work showed that energy measures can
correlate with shows (news programs in the Broadcast News corpus),
speakers, and so forth, rather than with the structural locations in
which we were interested.
Duration and pitch, on the
other hand, are relatively invariant to channel effects (to the extent that
they can be adequately extracted).  

In training, word boundaries were obtained from recognizer forced
alignments.  In testing on recognized words, we used alignments for
the 1-best recognition hypothesis.  Note that this results in a
mismatch between train and test data for the case of testing on
recognized words, that works {\em against} us. That is, the prosodic
models are trained on better alignments than can be expected in
testing; thus, the features selected may be suboptimal in the less
robust situation of recognized words. Therefore, we expect that any
benefit from the present, suboptimal approach would be only enhanced
if the prosodic models were based on recognizer alignments in training
as well.

\subsubsection{Features} \label{sec:features}

% NOTE: feat lists include the features used in the feature selection
% runs. Features used in any of the trees used in this paper
% are marked with a ``+''

We included features that, based on the descriptive literature, should
reflect breaks in the temporal and intonational contour. We developed
versions of such features that could be defined at each inter-word
boundary, and that could be extracted by completely automatic means,
without human labeling.  Furthermore, the features were designed to be
independent of word identities, for robustness to imperfect recognizer
output.

We began with a set of over 100 features, which, after initial
investigations, was pared down to a smaller set by eliminating features
that were clearly not at all useful (based on decision tree
experiments; see also Section~\ref{sec:feature-selection}). The
resulting set of features is described below. Features are grouped
into broad feature classes based on the kinds of measurements
involved, and the type of prosodic behavior they were designed to
capture.

        \paragraph{Pause features} \label{sec:pause-feats}

%        +PAU_DUR
%        +PREV_PAU_DUR
        
Important cues to boundaries between semantic units, such as sentences
or topics, are breaks in prosodic continuity, including pauses. We
extracted pause duration at each boundary based on recognizer output.
The pause model used by the recognizer was trained as an individual
phone, which during training could occur optionally between words.
In the case of no pause at the boundary, this pause duration feature
was output as 0. 

We also included the duration of the pause preceding the word before
the boundary, to reflect whether speech right before the boundary was
just starting up or continuous from previous speech.  Most inter-word
locations contained no pause, and were labeled as zero length.  We did
not need to distinguish between actual pauses and the short
segmental-related pauses (e.g., stop closures) inserted by the speech
recognizer, since models easily learned to distinguish the cases based
on duration.

We investigated both raw durations and durations normalized for pause
duration distributions from the particular speaker. Our models
selected the unnormalized feature over the normalized version,
possibly because of a lack of sufficient pause data per speaker. The
unnormalized measure was apparently sufficient to capture the gross
differences in pause duration distributions that separate boundary
from nonboundary locations, despite speaker variation within both
categories.
        
For the Broadcast News data, which contained mainly monologues and which
was recorded on a single channel, pause durations were undefined at
speaker changes.  For the Switchboard data there was significant
speaker overlap, and a high rate of backchannels (such as ``uh-huh'')
that were uttered by a listener during the speaker's turn.  Some of
these cases were associated with simultaneous speaker pausing and
listener backchanneling. Because the pauses here did not constitute
real turn boundaries, and because the Switchboard conversations were
recorded on separate channels, we included such speaker pauses in the
pause duration measure (i.e., even though a backchannel was uttered on
the other channel).

\paragraph{Phone and rhyme duration features} \label{sec:duration-feats}

%+RHYM_DUR_PH_bin
%+RHYM_NORM_DUR_PH_bin = NORM_RHYM_DUR/phones-in-coda
%AVG_PHONE_DUR_Z_bin
%AVG_VOWEL_DUR_Z_bin
%+MAX_PHONE_DUR_Z_bin
%+MAX_VOWEL_DUR_Z_bin
%RHYM_DUR_PH_ND_bin
%RHYM_NORM_DUR_PH_ND_bin
%VOW_DUR_Z_bin
%VOW_DUR_N_bin
        
Another well-known cue to boundaries in speech is a slowing down
toward the ends of units, or preboundary lengthening.  Preboundary
lengthening typically affects the nucleus and coda of syllables, so we
included measures here that reflected duration characteristics of the
last rhyme (nucleus plus coda) of the syllable preceding the boundary.

Each phone in the rhyme was normalized for inherent duration as
follows
\begin{equation} \label{eq:zscore}
\sum_{i}
\frac{phone\_dur_{i} - mean\_phone\_dur_{i}}{std\_dev\_phone\_dur_{i}}
\end{equation}
\noindent where $mean\_phone\_dur_{i}$ and $std\_dev\_phone\_dur_{i}$
are the mean and standard deviation of the current phone over all shows or
conversations in the training data.%
\footnote{Improvements in future work could include the use of
  triphone-based normalization (on a sufficiently large corpus to
  assure robust estimates), or of normalization based on syllable
  position and stress information (given a dictionary marked for this
  information).}  Rhyme features included the average normalized phone
duration in the rhyme, computed by dividing the measure in
Eq.~(\ref{eq:zscore}) by the number of phones in the rhyme, as well
as a variety of other methods for normalization.  To roughly capture
lengthening of prefinal syllables in a multisyllabic word, we also
recorded the longest normalized phone, as well as the longest
normalized vowel,
found in the preboundary word.%
\footnote{Using dictionary stress information would probably be a
  better approach. Nevertheless, one advantage of this simple method
  is a robustness to pronunciation variation, since the longest
  observed normalized phone duration is used, rather than some
  predetermined phone.}

We distinguished phones in filled pauses (such as ``um'' and ``uh'')
from those elsewhere, since it has been shown in previous work that
durations of such fillers (which are very frequent in Switchboard) are
considerably longer than those of spectrally similar vowels elsewhere
\cite{Shriberg:99}.  We also noted that for some phones, particularly
nasals, errors in the recognizer forced alignments in training
sometimes produced inordinately long (incorrect) phone durations. This
affected the robustness of our standard deviation estimates; to avoid
the problem we removed any clear outliers by inspecting the
phone-specific duration histograms prior to computing standard
deviations.

In addition to using phone-specific means and standard deviations over
all speakers in a corpus, we investigated the use of speaker-specific
values for normalization, backing off to cross-speaker values for
cases of low phone-by-speaker counts.  However, these features were
less useful than the features from data pooled over all speakers
(probably due to a lack of robustness in estimating the standard
deviations in the smaller, speaker-specific data sets).  Alternative
normalizations were also computed, including
$phone\_dur_{i}/mean\_phone\_dur_{i}$ (to avoid noisy estimates of
standard deviations), both for speaker-independent and
speaker-dependent means.

Interestingly, we found it necessary to bin the normalized duration
measures in order to reflect preboundary lengthening, rather than
segmental information.  Because these duration measures were
normalized by phone-specific values (means and standard deviations),
our decision trees were able to use certain specific feature values as
clues to word identities and, indirectly, to boundaries.
For example, the word ``I'' in the Switchboard
corpus is a strong cue to a sentence onset; normalizing by the
constant mean and standard deviation for that particular vowel
resulted in specific values that were ``learned'' by the models.  To
address this, we binned all duration features to remove the level of
precision associated with the phone-level correlations.

        \paragraph{F0 features} \label{f0-feats}

%Features reflecting F0 difference across boundary}
%*F0 difference across boundary
%requires same speaker
%11 feats
%+F0K_WRD_DIFF_HIHI_N
%F0K_WRD_DIFF_HILO_N
%+F0K_WRD_DIFF_LOLO_N
%F0K_WRD_DIFF_LOHI_N
%+F0K_WIN_DIFF_HIHI_N
%F0K_WIN_DIFF_HILO_N
%F0K_WIN_DIFF_LOLO_N
%F0K_WIN_DIFF_LOHI_N
%F0K_WRD_DIFF_MNMN_N
%F0K_WRD_DIFF_BEGBEG
%F0K_WRD_DIFF_ENDBEG

Pitch information is typically less robust and more difficult to model
than other prosodic features, such as duration. This is largely
attributable to variability in the way pitch is used across speakers
and speaking contexts, complexity in representing pitch patterns,
segmental effects, and pitch tracking discontinuities (such as
doubling errors and pitch halving, the latter of which is also
associated with nonmodal voicing). 

\begin{figure*}
\begin{center}
\strut\psfig{figure=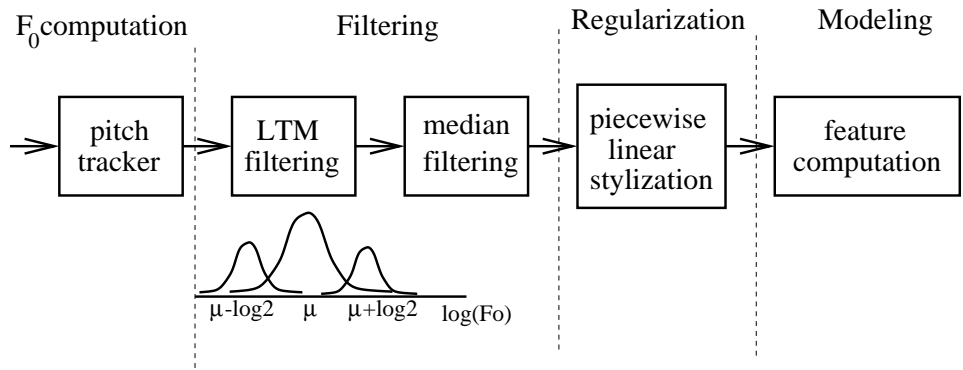,width=5in}
\end{center}
\caption{F0 processing}
\label{fig:f0-system}
\end{figure*}

To smooth out microintonation and tracking errors, simplify our F0
feature computation, and identify speaking-range parameters for each
speaker, we postprocessed the frame-level F0 output from a standard
pitch tracker.  We used an autocorrelation-based pitch tracker (the
``get\_f0'' function in ESPS/Waves \cite{ESPS},
with default parameter settings) to
generate estimates of frame-level F0 \cite{Talkin:95}.  Postprocessing
steps are outlined in Fig.~\ref{fig:f0-system} and are described
further in work on prosodic modeling for speaker verification
\cite{Sonmez:98}.

The raw pitch tracker output has two main noise sources, which are
minimized in the filtering stage.  F0 halving and doubling are
estimated by a lognormal tied mixture model (LTM) of F0, based on
histograms of F0 values collected from all data from the same speaker.%
\footnote{We settled on a cheating approach here, assuming speaker
  tracking information was available in testing, since automatic
  speaker segmentation and tracking was beyond the scope of this
  work.}  For the Broadcast News corpus we pooled data from the same
speaker over multiple news shows; for the Switchboard data, we
used only the data from one side of a conversation for each histogram.

For each speaker, the F0 distribution was modeled by three lognormal
modes spaced $\log 2$ apart in the log frequency domain.  The
locations of the modes were modeled with one tied parameter
($\mu-\log2$,$\mu$,$\mu+\log2$), variances were scaled to be the same
in the log domain, and mixture weights were estimated by an
expectation maximization (EM) algorithm.  This approach allowed
estimation of speaker F0 range parameters that proved useful for F0
normalization.

Prior to the regularization stage, median filtering smooths voicing
onsets during which the tracker is unstable, resulting in local
undershoot or overshoot.  We applied median filtering to windows of
voiced frames with a neighborhood size of 7 plus or minus 3 frames.
Next, in the regularization stage, F0 contours are fit by a simple
piecewise linear model
\[
\tilde{F_0} = \sum_{k=1}^K (a_k F_0 + b_k) I_{[x_{k-1} < F_0 \leq x_k]}
\]
where $K$ is the number of nodes, $x_k$ are the node locations, and
$a_k$ and $b_k$ are the linear parameters for a given region.  The
parameters are estimated by minimizing the mean squared error with a
greedy node placement algorithm. The smoothness of the fits is fixed
by two global parameters: the maximum mean squared error for deviation
from a line in a given region, and the minimum length of a region.

\begin{figure*}
\begin{center}
\strut\psfig{figure=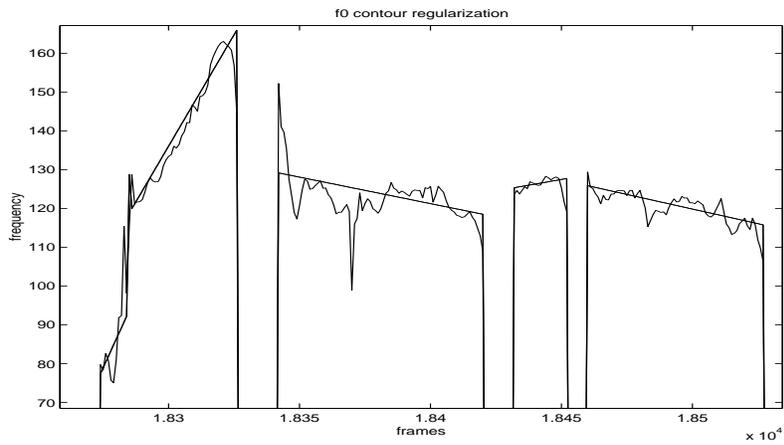,width=4.1in,height=2.3in}
\end{center}
\caption{F0 contour filtering and regularization}
\label{fig:f0-fit}
\end{figure*}

The resulting filtered and stylized F0 contour, an example of which is
shown in Fig.~\ref{fig:f0-fit}, enables robust extraction of
features such as the value of the F0 slope at a particular point, the
maximum or minimum stylized F0 within a region, and a simple
characterization of whether the F0 trajectory before a word boundary
is broken or continued into the next word.  In addition, over all data
from a particular speaker, statistics such as average slopes can be
computed for normalization purposes. These statistics, combined with
the speaker range values computed from the speaker histograms, allowed
us to easily and robustly compute a large number of F0 features, as
outlined in Section~\ref{sec:features}.  In exploratory work on
Switchboard, we found that the stylized F0 features yielded better
results than more complex features computed from the raw F0 tracks.
Thus, we restricted our input features to those computed from the
processed F0 tracks, and did the same for Broadcast News.

%*F0 of word on one side of boundary, relative to speakerspecific  range parameter
%20 features
%F0K_LR_LAST_KBASELN
%F0K_LR_WINMIN_KBASELN
%F0K_LR_MAXNEXT_KTOPLN
%F0K_LR_WINMAXNEXT_KTOPLN
%F0K_LR_MEANNEXT_KTOPLN
%+F0K_LR_MEAN_KBASELN
%F0K_DIFF_LAST_KBASELN
%F0K_DIFF_WINMIN_KBASELN
%F0K_DIFF_MAXNEXT_KTOPLN
%F0K_DIFF_WINMAXNEXT_KTOPLN
%F0K_DIFF_MEANNEXT_KTOPLN
%F0K_DIFF_MEAN_KBASELN
%F0K_MAXK_MODE_N
%F0K_MAXK_NEXT_MODE_N
%F0K_MAXK_MODE_Z
%F0K_MAXK_NEXT_MODE_Z
%F0K_ZRANGE_MEAN_KBASELN
%F0K_ZRANGE_MEAN_KTOPLN
%F0K_ZRANGE_MEANNEXT_KBASELN
%F0K_ZRANGE_MEANNEXT_KTOPLN
% (discrete in last feature case)
%5
%LAST_SLOPE
%LAST_SLOPE_N
%F0K_INWRD_DIFF (not slope but range covered within word)
%SLOPE_DIFF_N
%SLOPE_DIFF
%PATTERN_BOUNDARY (categorical)

\begin{figure}
\begin{center}
\strut\psfig{figure=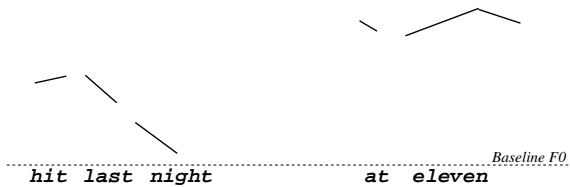,width=\columnwidth}
\end{center}
\caption{Schematic example of stylized F0 for voiced regions of the text. The speaker's estimated baseline F0 (from the lognormal tied mixture modeling) is also indicated.} 
\label{fig:f0-feats}
\end{figure}

We computed four different types of F0 features, all based on values
computed from the stylized processing, but each capturing a different
aspect of intonational behavior: (1) F0 {\em reset} features, (2) F0
{\em range} features, (3) F0 {\em slope} features, and (4) F0 {\em
  continuity} features.  The general characteristics captured can be
illustrated with the help of Fig.~\ref{fig:f0-feats}.

        {\em Reset features}. The first set of features was designed
        to capture the well-known tendency of speakers to reset pitch
        at the start of a new major unit, such as a topic or sentence
        boundary, relative to where they left off.  Typically the
        reset is preceded by a final fall in pitch associated with the
        ends of such units. Thus, at boundaries we expect a larger
        reset than at nonboundaries.  We took measurements from the
        stylized F0 contours for the voiced regions of the word
        preceding and of the word following the boundary.
        Measurements were taken at either the minimum, maximum, mean,
        starting, or ending stylized F0 value within the region
        associated with each of the words. Numerous features were
        computed to compare the previous to the following word; we
        computed both the log of the ratio between the two values, and
        the log of the difference between them, since it is unclear
        which measure would be better.  Thus, in
        Fig.~\ref{fig:f0-feats}, the F0 difference between ``at''
        and ``eleven'' would not imply a reset, but that between
        ``night'' and ``at'' would imply a large reset, particularly
        for the measure comparing the minimum F0 of ``night'' to the
        maximum F0 of ``at''. Parallel features were also computed
        based on the 200 ms windows rather than the words.

        {\em Range features}. The second set of features reflected the
        pitch range of a single word (or window), relative to one of
        the speaker-specific global F0 range parameters computed from
        the lognormal tied mixture modeling described earlier. We
        looked both before and after the boundary, but found features
        of the preboundary word or window to be the most useful for
        these tasks.  For the speaker-specific range parameters, we
        estimated F0 baselines, toplines, and some intermediate range
        measures.  By far the most useful value in our modeling was
        the F0 baseline, which we computed as occurring halfway
        between the first mode and the second mode in each
        speaker-specific F0 histogram, i.e., roughly at the bottom of
        the modal (nonhalved) speaking range.  We also estimated F0
        toplines and intermediate values in the range, but these
        parameters proved much less useful than the baselines across
        tasks.
        
        Unlike the reset features, which had to be defined as
        ``missing'' at boundaries containing a speaker change, the
        range features are defined at all boundaries for which F0
        estimates can be made (since they look only at one side of the
        boundary). Thus for example in Fig.~\ref{fig:f0-feats}, the
        F0 of the word ``night'' falls very close to the speaker's F0
        baseline, and can be utilized irrespective of whether or not
        the speaker changes before the next word.
        
        We were particularly interested in these features for the case
        of topic segmentation in Broadcast News, since due to the
        frequent speaker changes at actual topic boundaries we needed
        a measure that would be defined at such locations. We also
        expected speakers to be more likely to fall closer to the
        bottom of their pitch range for topic than for sentence
        boundaries, since the former implies a greater degree of
        finality.

        {\em Slope features}. Our final two sets of F0 features looked
        at the slopes of the stylized F0 segments, both for a word (or
        window) on only one side of the boundary, and for continuity
        across the boundary.  The aim was to capture local pitch
        variation such as the presence of pitch accents and boundary
        tones.  Slope features measured the degree of F0 excursion
        before or after the boundary (relative to the particular
        speaker's average excursion in the pitch range), or simply
        normalized by the pitch range on the particular word.
        
        {\em Continuity features}. Continuity features measured the
        change in slope across the boundary.  Here, we expected that
        continuous trajectories would correlate with nonboundaries,
        and broken trajectories would tend to indicate boundaries,
        regardless of difference in pitch values across words. For
        example, in Fig.~\ref{fig:f0-feats} the words ``last'' and
        ``night'' show a continuous pitch trajectory, so that it is
        highly unlikely there is a major syntactic or semantic
        boundary at that location.  We computed both scalar (slope
        difference) and categorical (rise-fall) features for inclusion
        in the experiments.

        \paragraph{Estimated voice quality features}

%PATTERN_WORD_HAS_H
%PATTERN_WORD_HAS_FINAL_H

Scalar F0 statistics (e.g., those contributing to slopes, or
minimum/maximum F0 within a word or region) were computed ignoring
any frames associated with F0 halving or doubling (frames whose
highest posterior was not that for the modal region).  However,
regions corresponding to F0 halving as estimated by the lognormal tied
mixture model showed high correlation with regions of creaky voice or
glottalization that had been independently hand-labeled by a
phonetician.  Since creak may correlate with our boundaries of
interest, we also included some categorical features, reflecting the
presence or absence of creak. 

We used two simple categorical features.  One feature reflected
whether or not pitch halving (as estimated by the model) was present
for at least a few frames, anywhere within the word preceding the
boundary. The second version looked at whether halving was present at
the end of that word.  As it turned out, while these two features
showed up in decision trees for some speakers, and in the patterns we
expected, glottalization and creak are highly speaker dependent and
thus were not helpful in our overall modeling.  However, for
speaker-dependent modeling, such features could potentially be more
useful.

\paragraph{Other features}
        
We included two types of nonprosodic features, turn-related features
and gender features.  Both kinds of features were legitimately
available for our modeling, in the sense that standard speech
recognition evaluations made this information known. Whether or not
speaker change markers would actually be available depends on the
application. It is not unreasonable however to assume this
information, since automatic algorithms have been developed for this
purpose (e.g., \opencite{NIST:euro99,LiuKubala:99,Sonmez:99}).  Such
nonprosodic features often interact with prosodic features.  For
example, turn boundaries cause certain prosodic features (such as F0
difference across the boundary) to be undefined, and speaker gender is
highly correlated with F0.  Thus, by including the features we could
better understand feature interactions and check for appropriateness
of normalization schemes.

Our turn-related features included whether or not the speaker changed
at a boundary, the time elapsed from the start of the turn, and the
turn count in the conversation. The last measure was included to
capture structure information about the data, such as the
preponderance of topic changes occurring early in Broadcast News
shows, due to short initial summaries of topics at the beginning of
certain shows.

%+TURN_F
%TURN_PAU_DUR pause dur at begin of turn
%+TURN_TIME time so far in turn
%TURN_TIME_N 
%TURN_CNT turn count in conversation

We included speaker gender mainly as a check to make sure the F0
processing was normalized properly for gender differences. That is, we
initially hoped that this feature would {\em not} show up in the
trees. However, we learned that there are reasons other than poor
normalization for gender to occur in the trees, including potential
truly stylistic differences between men and women, and structure
differences associated with gender (such as differences in lengths of
stories in Broadcast News). Thus, gender revealed some interesting
inherent interactions in our data, which are discussed further in
Section~\ref{sec:results-bn-topic}. In addition to speaker gender, we
included the gender of the listener, to investigate the degree to
which features distinguishing boundaries might be affected by
sociolinguistic variables.

%*gender
%+GEN
%GEN_PARTNER

\subsubsection{Decision trees} \label{sec:decision-trees}

As in past prosodic modeling work \cite{ShribergEtAl:eurospeech97}, we
chose to use CART-style decision trees \cite{Breiman:84}, as
implemented by the IND package \cite{IND}.  The software offers
options for handling missing feature values (important since we did
not have good pitch estimates for all data points), and is capable of
processing large amounts of training data.
Decision trees are probabilistic classifiers that can be characterized briefly
as follows.
Given a set of discrete or continuous features and a labeled training 
set, the decision tree construction algorithm repeatedly selects 
a single feature that, according to an information-theoretic criterion
(entropy),
has the highest predictive value for the classification task in question.%
\footnote{For multivalued or continuous features, the algorithm also
  determines optimal feature value subsets or thresholds,
  respectively, to compare the feature to.}  The feature queries are
arranged in a hierarchical fashion, yielding a tree of questions to be
asked of a given data point.  The leaves of the tree store
probabilities about the class distribution of all samples falling into
the corresponding region of the feature space, which then serve as
predictors for unseen test samples.  Various smoothing and pruning
techniques are commonly employed to avoid overfitting the model to the
training data.

Although any of several probabilistic classifiers (such as neural
networks, exponential models, or naive Bayes networks) could be used
as posterior probability estimators, decision trees allow us to add,
and automatically select, other (nonprosodic) features that might be
relevant to the task---including categorical features. Furthermore,
decision trees make no assumptions about the shape of feature
distributions; thus it is not necessary to convert feature values to
some standard scale. And perhaps most importantly, decision trees
offer the distinct advantage of interpretability. We have found that
human inspection of feature interactions in a decision tree fosters an
intuitive understanding of feature behaviors and the phenomena they
reflect.  This understanding is crucial for progress in developing
better features, as well as for debugging the feature extraction
process itself.

The decision tree served as a prosodic model for estimating the
posterior probability of a (sentence or topic) boundary at a given
inter-word boundary, based on the automatically extracted prosodic
features.  We define $F_i$ as the features extracted from a window
around the $i$th potential boundary, and $T_i$ as the boundary type
(boundary/no-boundary) at that position.  For each task, decision
trees were trained to predict the $i$th boundary type, i.e., to
estimate $P(T_i | F_i, W)$.  By design, this decision was only weakly
conditioned on the word sequence $W$, insofar as some of the prosodic
features depend on the phonetic alignment of the word models.
We preferred the weak conditioning for
robustness to word errors in speech recognizer output.
Missing feature values in $F_i$ occurred mainly for the F0 features
(due to lack of robust pitch estimates for an example), but also at
locations where features were inherently undefined (e.g., pauses at
turn boundaries). Such cases were handled in testing
by sending the test sample down
each tree branch with the proportion found in the training set at that
node, and then averaging the corresponding predictions.

\subsubsection{Feature selection algorithm} \label{sec:feature-selection}

Our initial feature sets contained a high degree of feature redundancy
because, for example, similar features arose from changing only
normalization schemes, and others (such as energy and F0) are
inherently correlated in speech production.  The greedy nature of the
decision tree learning algorithm implies that larger initial feature
sets can yield suboptimal results.  The availability of more features
provides greater opportunity for ``greedy'' features to be chosen;
such features minimize entropy locally but are suboptimal with respect
to entropy minimization over the whole tree.  Furthermore, it is
desirable to remove redundant features for computational efficiency
and to simplify interpretation of results.

To automatically reduce our large initial candidate feature set to an
optimal subset, we developed an iterative feature selection algorithm
that involved running multiple decision trees in training (sometimes
hundreds for each task).  The algorithm combines elements of
brute-force search with previously determined human-based heuristics
for narrowing the feature space to good groupings of features.
We used the entropy reduction of the overall tree after
cross-validation as a criterion for selecting the best subtree.
Entropy reduction is the difference in test-set entropy between the
prior class distribution and the posterior distribution estimated by
the tree. It is a more fine-grained metric than classification
accuracy, and is thus the more appropriate measure to use for any of
the model combination approaches described in
Section~\ref{sec:model-combination}.

The algorithm proceeds in two phases. In the first phase, the large
number of initial candidate features is reduced by a leave-one-out
procedure. Features that do not reduce performance when removed are
eliminated from further consideration.  The second phase begins with
the reduced number of features, and performs a beam search over all
possible subsets of features.  Because our initial feature set
contained over 100 features, we split the set into smaller subsets
based on our experience with feature behaviors.  For each subset we
included a set of ``core'' features, which we knew from human analyses
of results served as catalysts for other features.  For example, in
all subsets, pause duration was included, since without this feature
present, duration and pitch features are much less discriminative for
the boundaries of interest.\footnote{The success of this approach
  depends on the makeup of the initial feature sets, since highly
  correlated useful features can cancel each other out during the
  first phase.  This problem can be addressed by forming initial
  feature subsets that minimize within-set cross-feature
  correlations.}

\subsection{Language modeling}

The goal of language modeling for our segmentation tasks is to
capture information about segment boundaries contained in the word sequences.
We denote boundary classifications by 
$T = T_1, \ldots, T_K$ and use $W = W_1, \ldots, W_N$ for the word 
sequence.
Our general approach is to model the joint distribution of 
boundary types and words in a hidden Markov model (HMM), the hidden
variable in this case being the boundaries $T_i$ (or some related variable
from which $T_i$ can be inferred).
Because we had hand-labeled training data available for all 
tasks, the HMM parameters could be trained in supervised 
fashion.

The structure of the HMM is task specific, as described below,
but in all cases the Markovian character of the model allows us to
efficiently perform the probabilistic inferences desired.
For example, for topic segmentation we extract the most likely {\em overall}
boundary classification
\begin{equation}
        \argmax_T P(T | W) \quad ,
\end{equation}
using the Viterbi algorithm \cite{Viterbi:67}.
This optimization criterion is appropriate because the topic segmentation
evaluation metric 
prescribed by the TDT program \cite{TDT2} rewards overall consistency of
the segmentation.\footnote{
For example, given three sentences $s_1 s_2 s_3$ and strong evidence
that there is a topic boundary between $s_1$ and $s_3$, it is better
to output a boundary either before or after $s_2$, but not in both places.}

For sentence segmentation, the evaluation metric simply counts the number
of correctly labeled boundaries (see Section~\ref{sec:metrics}).
Therefore, it is advantageous
to use the slightly more complex forward-backward algorithm \cite{Baum:70}
to maximize the posterior probability of each individual boundary
classification $T_i$
\begin{equation}
        \argmax_{T_i} P(T_i | W) \quad .
\end{equation}
This approach minimizes the expected per-boundary classification error rate
\cite{Dermatas:95}.

\subsubsection{Sentence segmentation}

We relied on a hidden-event N-gram language
model (LM) \cite{StoShr:icslp96,SOBS:icslp98}.
The states of
the HMM consist of the end-of-sentence status of each word (boundary
or no-boundary), plus any preceding words and possibly boundary tags to
fill up the N-gram context ($N = 4$ in our experiments).  Transition
probabilities are given by N-gram probabilities estimated from
annotated, boundary-tagged training data using Katz backoff \cite{Katz:87}.
For example, the bigram parameter $P(\mbox{\tt <S>}| \mbox{\tt tonight})$
gives the probability of a sentence boundary following the word ``tonight''.
HMM observations consist of only the current word portion of the 
underlying N-gram state (with emission likelihood 1), constraining
the state sequence to be consistent with the observed word sequence.

\subsubsection{Topic segmentation}

We first constructed 100 individual unigram topic cluster language
models, using the multipass {\em k}-means algorithm described in
\cite{Yamron:98}. We used the pooled Topic Detection and Tracking
(TDT) Pilot and TDT-2 training data \cite{TDT2:darpa99}.  We removed
stories with fewer than 300 and more than 3000 words, leaving 19,916
stories with an average length of 538 words. Then, similar to the
Dragon topic segmentation approach \cite{Yamron:98}, we built an HMM
in which the states are topic clusters, and the observations are
sentences. The resulting HMM forms a complete graph, allowing
transition between any two topic clusters.  In addition to the basic
HMM segmenter, we incorporated two states for modeling the initial and
final sentences of a topic segment.  We reasoned that this can capture
formulaic speech patterns used by broadcast speakers.  Likelihoods for
the start and end models are obtained as the unigram language model
probabilities of the topic-initial and final sentences, respectively,
in the training data.  Note that single start and end states are
shared for all topics, and traversal of the initial and final states
is optional in the HMM topology.  The topic cluster models work best
if whole blocks of words or ``pseudo-sentences'' are evaluated against
the topic language models (the likelihoods are otherwise too noisy).
We therefore presegment the data stream at pauses exceeding 0.65
second, as process we will refer to as ``chopping''.

\subsection{Model combination} \label{sec:model-combination}

We expect prosodic and lexical segmentation cues to be partly
complementary, so that combining both knowledge sources should give
superior accuracy over using each source alone.  This raises the issue
of how the knowledge sources should be integrated.  Here, we describe
two approaches to model combination that allow the component prosodic
and lexical models to be retained without much modification.  While
this is convenient and computationally efficient, it prevents us from
explicitly modeling interactions (i.e., statistical dependence)
between the two knowledge sources.  Other researchers have proposed
model architectures based on decision trees \cite{HeemanAllen:97} or
exponential models \cite{Beeferman:99} that can potentially integrate
the prosodic and lexical cues discussed here.  In other work
\cite{SOBS:icslp98,TurEtAl:CL2000} we have started to study
integrated approaches for the segmentation tasks studied here,
although preliminary results show that the simple combination
techniques are very competitive in practice.

\subsubsection{Posterior probability interpolation}
        \label{sec:interpolate-posteriors}

Both the prosodic decision tree and the language model
(via the forward-backward algorithm) estimate posterior probabilities
for each boundary type $T_i$.
We can arrive at a better posterior estimator by linear interpolation:
\begin{equation}
        P(T_i | W , F) \approx  \lambda P_{\rm LM}(T_i | W)
                                + (1 - \lambda) P_{\rm DT}(T_i | F_i, W)
\end{equation}
where $\lambda$ is a parameter optimized on held-out data
to optimize the overall model performance.

\subsubsection{Integrated hidden Markov modeling}

Our second model combination approach is based on the idea that 
the HMM used for lexical modeling can be extended to ``emit'' both
words and prosodic observations.  The goal is to obtain an HMM
that models the joint distribution $P(W,F,T)$ of word sequences $W$, prosodic 
features $F$, and hidden boundary types $T$ in a Markov model.
With suitable independence assumptions we can then
apply the familiar HMM techniques to compute 
\[
        \argmax_{T} P(T | W, F)
\]
or 
\[
        \argmax_{T_i} P(T_i | W, F) \quad ,
\]
which are now conditioned on both lexical and prosodic cues.
We describe this approach for sentence segmentation HMMs; the
treatment for topic segmentation HMMs is mostly analogous but somewhat
more involved, and described in detail elsewhere \cite{TurEtAl:CL2000}.

To incorporate the prosodic information into the HMM, we model
prosodic features as emissions from relevant HMM states, with
likelihoods $P(F_i | T_i, W)$, where $F_i$ is the feature vector pertaining
to potential boundary $T_i$.     For example, an HMM state representing 
a sentence boundary {\tt <S>} at the current position would be 
penalized with the likelihood $P(F_i | \mbox{\tt <S>})$.
We do so based on the assumption that
prosodic observations are conditionally independent of each other
given the boundary types $T_i$ and the words $W$.
Under these assumptions, a complete path through the HMM is associated 
with the total probability
\begin{equation}
        P(W, T) \prod_{i} P(F_i | T_i, W) = P(W, F, T) \quad,
\end{equation}
as desired.

The remaining problem is to estimate the likelihoods $P(F_i | T_i, W)$.
Note that the decision tree estimates posteriors $P_{\rm DT}(T_i|F_i,W)$.
These can be converted to likelihoods using Bayes' rule as in
\begin{equation}
P(F_i | T_i, W) = { P(F_i|W) P_{\rm DT}(T_i|F_i,W) \over P(T_i|W) } \quad .
\end{equation}
The term $ P(F_i|W)$ is a constant for all choices of $T_i$ and can thus
be ignored when choosing the most probable one.
Next, because our prosodic model is purposely not conditioned on 
word identities, but only on aspects of $W$ that relate to time alignment,
we approximate $P(T_i|W) \approx P(T_i)$.
Instead of explicitly dividing the posteriors, we
prefer to downsample the training set to make
$P(T_i = {\bf yes}) = P(T_i = {\bf no}) = {1 \over 2}$.
A beneficial side effect of this approach is that
the decision tree models the lower-frequency events (segment boundaries)
in greater detail than if presented with the raw, highly skewed class
distribution.
 
When combining probabilistic models of different
types, it is advantageous to weight the contributions of the
language models and the prosodic trees relative to each other.
We do so by introducing a tunable {\em model combination weight} (MCW),
and by using $P_{\rm DT}(F_i|T_i, W)^{\rm MCW}$ as the effective prosodic
likelihoods.  The value of MCW is optimized on held-out data.

\subsubsection{HMM posteriors as decision tree features}

A third approach could be used to combine the language and prosodic
models, although for practical reasons we chose not to use it in this work.
In this approach, an HMM incorporating only lexical
information is used to compute posterior probabilties of boundary
types, as described in Section~\ref{sec:interpolate-posteriors}.  A
prosodic decision tree is then trained, using the HMM posteriors as
additional input features.  The tree is free to combine the word-based
posteriors with prosodic features; it can thus model limited forms of
dependence between prosodic and word-based information (as summarized
in the posteriors).

A severe drawback of using posteriors in the decision tree, however,
is that in our current paradigm, the HMM is trained on correct words.
In testing, the tree may therefore grossly overestimate the
informativeness of the word-based posteriors based on automatic
transcriptions.  Indeed, we found that on a hidden-event detection
task similar to sentence segmentation \cite{SOBS:icslp98} this model
combination method worked well on true words, but faired worse than
the other approaches on recognized words.
To remedy the mismatch between training and testing of the combined
model, we would have to train, as well as test, on recognized words;
this would require computationally intensive processing of a large
corpus.  For these reasons, we decided not to use HMM posteriors as
tree features in the present studies.

\subsubsection{Alternative models}

A few additional comments are in order regarding our choice of model
architectures and possible alternatives.  The HMMs used for lexical
modeling are {\em likelihood models}, i.e., they model the
probabilities of observations given the hidden variables (boundary
types) to be inferred, while making assumptions about the independence
of the observations given the hidden events.  The main virtue of HMMs
in our context is that they integrate the local evidence (words and
prosodic features) with models of context (the N-gram history) in a
very computationally efficient way (for both training and testing).  A
drawback is that the independence assumptions may be inappropriate and
may therefore inherently limit the performance of the model.

The decision trees used for prosodic modeling, on the other hand, are
{\em posterior models}, i.e., they directly model the probabilities of
the unknown variables given the observations.  Unlike 
likelihood-based models, this has the advantages that model training
explicitly enhances discrimination between the target classifications
(i.e., boundary types), and that input features can be combined easily
to model interactions between them.  Drawbacks are the sensitivity to
skewed class distributions (as pointed out in the previous section),
and the fact that it becomes computationally expensive to model interactions
between multiple target variables (e.g., adjacent boundaries).
Furthermore, input features with large discrete ranges (such as the set
of words) present practical problems for many posterior model
architectures.

Even for the tasks discussed here, other modeling choices would have
been practical, and await comparative study in future work.  For
example, posterior lexical models (such as decision trees or neural
network classifiers) could be used to predict the boundary types from
words and prosodic features together, using word-coding techniques
developed for tree-based language models \cite{Bahl:89}.  Conversely,
we could have used prosodic likelihood models, removing the need to
convert posteriors to likelihoods.  For example, the continuous
feature distributions could be modeled with (mixtures of)
multidimensional Gaussians (or other types of distributions), as is
commonly done for the spectral features in speech recognizers
(\opencite{Digalakis:94}, among others).

\subsection{Data}
\subsubsection{Speech data and annotations}

Switchboard data used in sentence segmentation was drawn from a subset
of the corpus \cite{SWBD} that had been hand-labeled for sentence
boundaries \cite{SWBD-DF} by the Linguistic Data Consortium (LDC).
Broadcast News data for topic and sentence segmentation was extracted
from the LDC's 1997 Broadcast News (BN) release.  Sentence boundaries
in BN were automatically determined using the MITRE sentence tagger
\cite{Palmer:97} based on capitalization and punctuation in the
transcripts.  Topic boundaries were derived from the SGML markup of
story units in the transcripts.  Training of Broadcast News language
models for sentence segmentation also used an additional 130 million
words of text-only transcripts from the 1996 Hub-4 language model
corpus, in which sentence boundaries had been marked by SGML tags.

\subsubsection{Training, tuning, and test sets}

Table~\ref{tab:data} shows the amount of data used for the 
various tasks.  For each task, separate datasets were used for
model training, for tuning any free parameters (such as the model combination
and posterior interpolation weights), and for final testing.  
In most cases the language model and the prosodic model components used
different amounts of training data.

\begin{table*}
\caption{\label{tab:data}
        Size of speech data sets used for model training and testing
	for the three segmentation tasks}
\begin{center}
\begin{tabular}{lllll}
\hline
Task            & \multicolumn{2}{l}{Training} & Tuning        & Test \\
		\cline{2-3}
                & LM            & Prosody       &               &       \\
\hline
SWB Sentence    %       di97-trn                 di97-hld-out-1  di97-hld-out-2
                & 1788 sides	& 1788 sides	& 209 sides    & 209 sides \\
(transcribed)   & (1.2M words)	& (1.2M words)	& (103K words)& (101K words)\\
\hline
SWB Sentence    %                                 nbest-train     ws97-devtest
                & 1788 sides	& 1788 sides	& 12 sides     & 38 sides \\
(recognized)    & (1.2M words)	& (1.2M words)	& (6K words) & (18K words) \\
\hline
BN Sentence     % train+hldout    train           devtest-1       evaltest
                & 103 shows + BN96 & 93 shows   & 5 shows       & 5 shows \\
                & (130M words)  & (700K words)  & (24K words)   & (21K words) \\
\hline
BN Topic        %                train            hldout          devtest
                & TDT + TDT2    & 93 shows      & 10 shows      & 6 shows \\
                & (10.7M words) & (700K words)  & (205K words)  & (44K words) \\
\hline
\end{tabular}
\end{center}
\end{table*}

As is common for speech recognition evaluations on Broadcast News,
frequent speakers (such as news anchors) appear in both training and
test sets. By contrast, in Switchboard our train and test sets did not
share any speakers. In both corpora, the average word count per
speaker decreased roughly monotonically with the percentage of
speakers included. In particular, the Broadcast News data contained a
large number of speakers who contributed very few words.  A reasonably
meaningful statistic to report for words per speaker is thus a
weighted average, or the average number of datapoints by the same
speaker. On that measure, the two corpora had similar statistics:
6687.11 and 7525.67 for Broadcast News and Switchboard, respectively.
% other stats: 
%               SWB             BN
% mean          3884.52         588.15
% stdev         3766.03         1894.69
% min           205             1
% max           24464           30901
% 

\subsubsection{Word recognition}

Experiments involving recognized words used the 1-best output from
SRI's  DECIPHER  large-vocabulary  speech recognizer.
We simplified processing by skipping several of the computationally expensive
or cumbersome steps often used for optimum performance, such as 
acoustic adaptation and multiple-pass decoding.  The recognizer
performed one bigram decoding pass, followed by a single N-best rescoring
pass using a higher-order language model.
The Switchboard test set was decoded with a word error rate of 
46.7\% using acoustic models developed for the 1997 Hub-5 evaluation
\cite{LVCSR:97}.  The Broadcast News recognizer was based on the 
1997 SRI Hub-4 recognizer \cite{Sankar:darpa98} and had a word error
rate of 30.5\% on the test set used in our study.

\subsubsection{Evaluation metrics}
        \label{sec:metrics}

Sentence segmentation performance for true words was measured by
boundary classification error, i.e. the percentage of word boundaries
labeled with the incorrect class. For recognized words, we first
performed a string alignment of the automatically labeled recognition
hypothesis with the reference word string (and its segmentation).
Based on this alignment we then counted the number of incorrectly
labeled, deleted, and inserted word boundaries, expressed as a
percentage of the total number of word boundaries. This metric yields
the same result as the boundary classification error rate if the word
hypothesis is correct.  Otherwise, it includes additional errors from
inserted or deleted boundaries, in a manner similar to standard word
error scoring in speech recognition.  Topic segmentation was evaluated
using the metric defined by NIST for the TDT-2 evaluation \cite{TDT2}.

\section{Results and discussion} \label{sec:results-discussion}

The following sections describe results from the prosodic modeling
approach,  for each of our three tasks.  The first three sections focus on 
the tasks individually, detailing the features used in the best-performing
tree.  For sentence segmentation, we report on trees trained on
non-downsampled data, as used in the posterior interpolation approach.
For all tasks, including topic segmentation, we also trained
downsampled trees for the HMM combination approach.  Where both types
of trees were used (sentence segmentation), feature usage on
downsampled trees was roughly similar to that of the non-downsampled
trees, so we describe only the non-downsampled trees. For topic
segmentation, the description refers to a downsampled tree.

In each case we then look at results from combining the prosodic
information with language model information, for both transcribed and
recognized words. Where possible (i.e., in the sentence segmentation
tasks), we compare results for the two alternative model integration
approaches (combined HMM and interpolation).  In the next two
sections, we compare results across both tasks and speech corpora.  We
discuss differences in which types of features are helpful for a task,
as well as differences in the relative reduction in error achieved by
the different models, using a measure that tries to normalize for the
inherent difficulty of each task. Finally, we discuss issues for
future work.

\subsection{Task 1: Sentence segmentation of Broadcast News data}

\subsubsection{Prosodic feature usage}

%                               *202as
%                                (featsfrom201)(better than dhfeats)
%                                202as PAU_DUR              0.46058 
%                                202as TURN_F               0.41821 
%                                202as F0K_WRD_DIFF_LOLO_N  0.07992 
%                                202as F0K_WRD_DIFF_HIHI_N  0.02162 
%                                202as CODA_DUR_PH_bin      0.01080 
%                                202as F0K_WIN_DIFF_HIHI_N  0.00887 

The best-performing tree identified six features for this task, which
fall into four groups.  To summarize the relative importance of the
features in the decision tree we use a measure we call ``feature
usage'', which is computed as the relative frequency with which that
feature or feature class is queried in the decision tree. The measure
increments for each sample classified using that feature; 
features used higher in the tree classify more samples and therefore have
higher usage values. The feature usage was as follows
(by type of feature):

\begin{itemize}
\item (46\%) Pause duration at boundary
\item (42\%) Turn/no turn at boundary
\item (11\%) F0 difference across boundary
\item (01\%) Rhyme duration
\end{itemize}

\begin{figure*}
\begin{center}
\hspace{2in}\strut\psfig{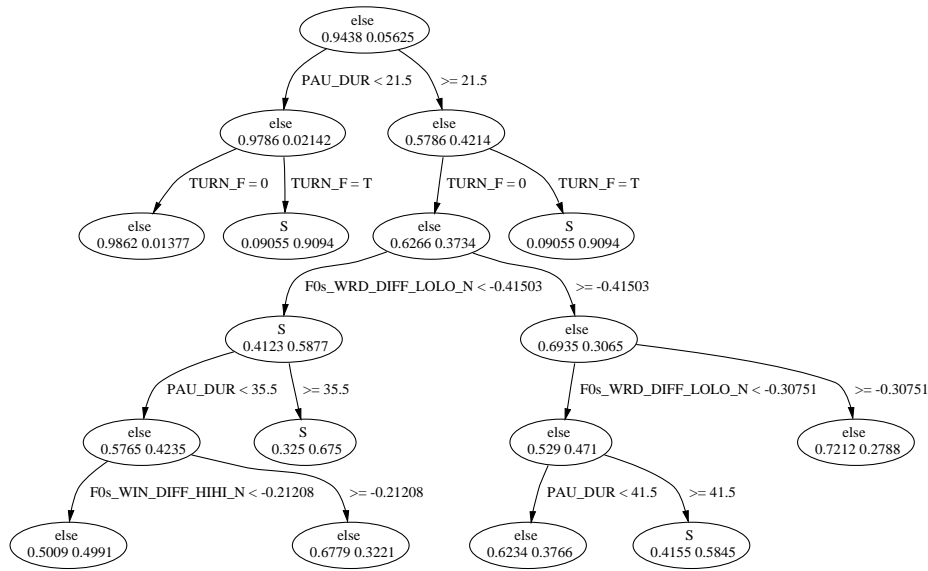}
\end{center}
\caption{Top levels of decision tree selected for the Broadcast News sentence segmentation task. Nodes contain the percentage of ``else'' and ``S'' (sentence) boundaries, respectively, and are labeled with the majority class. PAU\_DUR=pause duration, F0s=stylized F0 feature reflecting ratio of speech before the boundary to that after that boundary, in the log domain.}
\label{fig:tree-sent-bn} % tree is 202as (nondownsampled)
\end{figure*}

The main features queried were pause, turn, and F0. To
understand whether they behaved in the manner expected based on the
descriptive literature, we inspected the decision tree. The tree for
this task had 29 leaves; we show the top portion of it in
Fig.~\ref{fig:tree-sent-bn}.

The behavior of the features  is precisely that expected from the
literature.  Longer pause durations at the boundary imply a higher
probability of a sentence boundary at that location. Speakers exchange
turns almost exclusively at sentence boundaries in this corpus, so the
presence of a turn boundary implies a sentence boundary. The F0
features all behave in the same way, with lower negative values
raising the probability of a sentence boundary. These features reflect
the log of the ratio of F0 measured within the word (or window) preceding the boundary to
the F0 in the word (or window) after the boundary. Thus, lower negative values imply
a larger pitch reset at the boundary, consistent with what we would expect.

\subsubsection{Error reduction from prosody}

Table~\ref{tab:bn-sent-results} summarizes the results on both
transcribed and recognized words, for various sentence segmentation
models for this corpus.  The baseline (or ``chance'') performance for
true words in this task is 6.2\% error, obtained by labeling all
locations as nonboundaries (the most frequent class). For recognized
words, it is considerably higher; this is due to the non-zero lower
bound resulting if one accounts for locations in which the 1-best
hypothesis boundaries do not coincide with those of the reference
alignment. ``Lower bound'' gives the lowest segmentation error rate possible
given the word boundary mismatches due to recognition errors.

\begin{table*}
\caption{\label{tab:bn-sent-results}
        Results for sentence segmentation on Broadcast News}
\begin{center}
\begin{tabular}{lll}
\hline
Model           & Transcribed words     & Recognized words \\
\hline
LM only (130M words)      &     4.1             & 11.8          \\
Prosody only (700K words) &     3.6             & 10.9          \\
Interpolated              &     3.5             & 10.8          \\
Combined HMM              &     3.3             & 11.7          \\
			&			& \\
Chance                    &     6.2             & 13.3          \\
Lower bound               &     0.0             & 7.9           \\
\hline
\end{tabular}
\\ Values are word
boundary classification error rates (in percent).
\end{center}
\end{table*}
Results show that the prosodic model alone performs better
than a word-based language model, despite the fact that the language model
was trained on a much larger data set.  Furthermore, the prosodic
model is somewhat more robust to errorful recognizer output than the
language model, as measured by the absolute increase in error rate in
each case.  Most importantly, a statistically significant error
reduction is achieved by combining the prosodic features with the
lexical features, for both integration methods. The relative error
reduction is 19\% for true words, and 8.5\% for recognized words.
This is true even though both models contained turn information, thus
violating the independence assumption made in the model combination.

\subsubsection{Performance without F0 features} % NEW

A question one may ask in using the prosody features, is how the model
would perform without any F0 features. Unlike pause, turn, and
duration information, the F0 features used are not typically extracted
or computed in most ASR systems.  We ran comparison experiments on
all conditions, but removing all F0 features from the input to the
feature selection algorithm.  Results are shown in
Table~\ref{tab:bn-sent-results-nof0}, along with the previous results
using all features, for comparison.

% full results set:
%                       All features    No F0 feats     no F0, duration feats
%TRUE WORDS
%Prosody tree only      3.55            3.80            3.74
%+700K LM               3.79(*)         3.61            3.61
%+130M LM               3.34            3.20            3.17
%RECOGNIZED WORDS
%Prosody tree only      10.90           11.33           11.16
%+700K LM               10.89           11.24           11.29(*)
%+130M LM               10.82           11.14           11.07
%(*) combined model did better on development set, but worse than tree
%       baseline on independent test set.
%Note1: only 130M LM results appear in the result table in the paper,
%  so your could ignore the 700K LM results.
%Note2: combination method for true words was combined HMM,
%   interpolation of posteriors for recognized words.
% report only 130M LM and the All/NoF0 columns in paper, 

\begin{table*}
\caption{\label{tab:bn-sent-results-nof0} 
       Results for sentence segmentation on Broadcast News, with and
	without F0 features}
\begin{center}
\begin{tabular}{lll}
\hline
Model                      & Transcribed Words     & Recognized Words \\
\hline
LM only (130M words)      &     4.1         & 11.8     \\
			&		& \\
All Prosody Features: & & \\
~~~Prosody only (700K words)  &    3.6         &  10.9     \\
~~~Prosody+LM: Combined HMM   &    3.3         &           \\
~~~Prosody+LM: Interpolation  &                &  10.8     \\ 
				& 	& \\
No F0 Features: & & \\
~~~Prosody only (700K words)  &   3.8          &  11.3     \\
~~~Prosody+LM: Combined HMM   &   3.2          &          \\
~~~Prosody+LM: Interpolation  &                &  11.1     \\ 
			& 	& \\
Chance                    &     6.2         & 13.3       \\
Lower bound               &     0.0         & 7.9        \\
\hline
\end{tabular}
\\
       Values are word boundary classification error rates (in percent). 
       For the integrated (``Prosody + LM'') models, results are given for the
	optimal model only 
       (combined HMM for true words, interpolation of posteriors for recognized words.)
\end{center}
\end{table*}

As shown, the effect of removing F0 features reduces model accuracy
for prosody alone, for both true and recognized words.  In the case of
the true words, model integration using the no-F0 prosodic tree
actually fares slightly better than that which used all features,
despite similar model combination weights in the two cases.  The
effect is only marginally significant in a Sign test, so it may
indicate chance variation. However it could also indicate a higher
degree of correlation between true words and the prosodic features
that indicate boundaries, when F0 is included.  However, for
recognized words, the model with all prosodic features is superior to
that without the F0 features, both alone and after integration with
the language model.

\subsection{Task 2: Sentence segmentation of Switchboard data}

\subsubsection{Prosodic feature usage}

%                                *054as
%                                (101dh70feats)
%                                054as MAX_VOWEL_DUR_Z_bin  0.26299 
%                                054as CODA_DUR_PH_bin      0.18694 
%                                054as PAU_DUR              0.18271 
%                                054as TURN_F               0.16573 
%                                054as PREV_PAU_DUR         0.15097 
%                                054as MAX_PHONE_DUR_Z_bin  0.03257 
%                                054as TURN_TIME            0.00995 
%                                054as CODA_NORM_DUR_PH_bin 0.00814 

%                                .49 dur
%                                .18 pause
%                                .15 pre pau dur
%                                .17 turn
%                                .01 turntime

Switchboard sentence segmentation made use of a markedly different distribution of
features than observed for Broadcast News. For Switchboard, the best-performing
tree found by the feature selection algorithm had a feature usage as follows:

\begin{itemize}
\item (49\%) Phone and rhyme duration preceding boundary
\item (18\%) Pause duration at boundary
\item (17\%) Turn/no turn at boundary
\item (15\%) Pause duration at {\em previous} word boundary
\item (01\%) Time elapsed in turn
\end{itemize}

\noindent Clearly, the primary feature type used here is preboundary duration,
a measure that was used only a scant 1\% of the time for the same task
in news speech.  Pause duration at the boundary was also useful, but
not to the degree found for Broadcast News.

Of course, it should be noted in comparing feature usage across
corpora and tasks that results here pertain to comparisons of {\em the
  most parsimonious, best-performing model} for each corpus and task.
That is, we do not mean to imply that an individual feature such as
preboundary duration is not useful in Broadcast News, but rather that
the minimal and most successful model for that corpus makes little use
of that feature (because it can make better use of other features).
Thus, it cannot be inferred from these results that some feature not
heavily used in the minimal model is not helpful. The feature may be
useful on its own; however, it is not as useful as some other
feature(s) made
available in this study.%
\footnote{One might propose a more thorough investigation by reporting
  performance for one feature at a time. However, we found in
  examining such results that typically our features required the
  presence of one or more additional features in order to be helpful.
  (For example, pitch features required the presence of the pause
  feature.) Given the large number of features used, the number of
  potential combinations becomes too large to report on fully here.}

\begin{figure*}
\begin{center}
\strut\psfig{figure=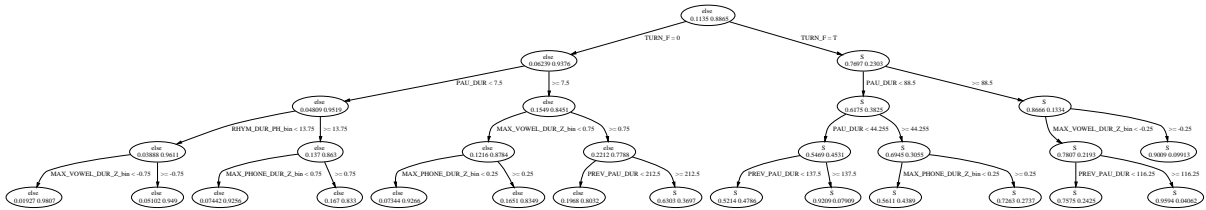,width=\textwidth,angle=-90}
\end{center}
\caption{Top levels of decision tree selected for the Switchboard sentence segmentation task. Nodes contain the percentage of  ``S'' (sentence) and ``else''  boundaries, respectively, and are labeled with the majority class. ``PAU\_DUR''=pause duration, ``RHYM''=syllable rhyme. VOWEL, PHONE and RHYME features apply to the word before the boundary.}
\label{fig:tree-sent-sb} 
\end{figure*}

The two ``pause'' features are not grouped together, because they
represent fundamentally different phenomena. The second pause feature
essentially captured the boundaries after one word such as ``uh-huh''
and ``yeah'', which for this work had been marked as followed by
sentence
boundaries (``yeah {\tt <Sent>} i know what you mean'').%
\footnote{``Utterance'' boundary is probably a better term, but for
  consistency we use the term ``sentence'' boundary for these dialogue
  act boundaries as well.} The previous pause in this case was time
that the speaker had spent in listening to the other speaker (channels
were recorded separately and recordings were continuous on both
sides). Since one-word backchannels (acknowledgments such as
``uh-huh'') and other short dialogue acts make up a large percentage of
sentence boundaries in this corpus, the feature is used fairly often.
The turn features also capture similar phenomena related to
turn-taking.  The leaf count for this tree was 236, so we display only
the top portion of the tree in Fig.~\ref{fig:tree-sent-sb}.

Pause and turn information, as expected, suggested sentence
boundaries.  Most interesting about this tree was the consistent behavior
of duration features, which gave higher probability to a sentence
boundary when lengthening of phones or rhymes was detected in the word
preceding the boundary.  Although this is in line with descriptive
studies of prosody, it was rather remarkable to us that duration would
work at all, given the casual style and speaker variation in
this corpus, as well as the somewhat noisy forced alignments for the
prosodic model training.

\subsubsection{Error reduction from prosody}

\begin{table}
\caption{\label{tab:swb-sent-results}
        Results for sentence segmentation on Switchboard}
\begin{center}
\begin{tabular}{lll}
\hline
Model           & Transcribed     	& Recognized \\
		& words			& words \\
\hline
LM only         &       4.3             & 22.8          \\
Prosody only    &       6.7             & 22.9          \\
Interpolated    &       4.1             & 22.2          \\
Combined HMM    &       4.0             & 22.5          \\
		& 			& \\
Chance          &       11.0            & 25.8          \\
Lower bound     &       0.0             & 17.6          \\
\hline
\end{tabular}
\\ Values are word boundary classification error rates (in percent).
\end{center}
\end{table}

Unlike the previous results for the same task on Broadcast News, we
see in Table~\ref{tab:swb-sent-results} that for Switchboard data,
prosody alone is not a particularly good model. For transcribed words
it is considerably worse than the language model; however, this
difference is reduced for the case of recognized words (where the
prosody shows less degradation than the language model).

Yet, despite the poor performance of prosody alone, combining prosody
with the language model resulted in a statistically significant
improvement over the language model alone (7.0\% and 2.6\% relative
for true and recognized words, respectively). All differences were
statistically significant, including the difference in performance
between the two model integration approaches.  Furthermore, the
pattern of results for model combination approaches observed for
Broadcast News holds as well: the combined HMM is superior for the
case of transcribed words, but suffers more than the interpolation
approach when applied to recognized words.

\subsection{Task 3: Topic segmentation of Broadcast News data}
        \label{sec:results-bn-topic}

\subsubsection{Prosodic feature usage}

The feature selection algorithm determined five feature types most
helpful for this task:

%  106dh-52 PAU_DUR              0.42649 
%  106dh-52 F0K_LR_MEAN_KBASELN  0.25380 
%  106dh-52 F0K_WRD_DIFF_MNMN_NG 0.10538 
%  106dh-52 TURN_F               0.09339 
%   106dh-52 GEN                  0.06826 
%  106dh-52 TURN_TIME            0.05267 

%  .43 pause
%  .36 f0
%  .09 turn
%  .07 gen
%  .05 turntime 

\begin{itemize}
\item (43\%) Pause duration at boundary
\item (36\%) F0 range
\item (09\%) Turn/no turn at boundary
\item (07\%) Speaker gender
\item (05\%) Time elapsed in turn
\end{itemize}

The results are somewhat similar to those seen earlier for
sentence segmentation in Broadcast News, in that pause, turn, and F0
information are the top features. However, the feature usage here
differs considerably from that for the sentence segmentation task, in
that here we see a much higher use of F0 information. 

Furthermore, the most important F0 feature was a range feature
(log ratio of the preceding word's F0 to the speaker's F0 baseline), which
was used 2.5 times more often in the tree than the F0 feature based on
difference across the boundary.  The range feature does not require
information about F0 on the other side of the boundary; thus, it could
be applied regardless of whether there was a speaker change at
that location.  This was a much more important issue for topic
segmentation than for sentence segmentation, since the percentage of
speaker changes is higher in the former than in the latter.

It should be noted, however, that the importance of pause duration is
underestimated.  As explained earlier, pause duration was also
used {\em prior} to tree building, in the chopping process.  The
decision tree was applied only to boundaries exceeding a certain
duration.  Since the duration threshold was found by 
optimizing for the TDT error criterion, which assigns greater weight
to false alarms than to false rejections, the resulting pause
threshold is quite high (over half a second).  Separate experiments
using boundaries below our chopping threshold show that trees
distinguish much shorter pause durations for segmentation decisions,
implying that prosody could potentially yield an even larger relative
advantage for error metrics favoring a shorter chopping threshold.

\begin{figure*}
\begin{center}
\strut\psfig{figure=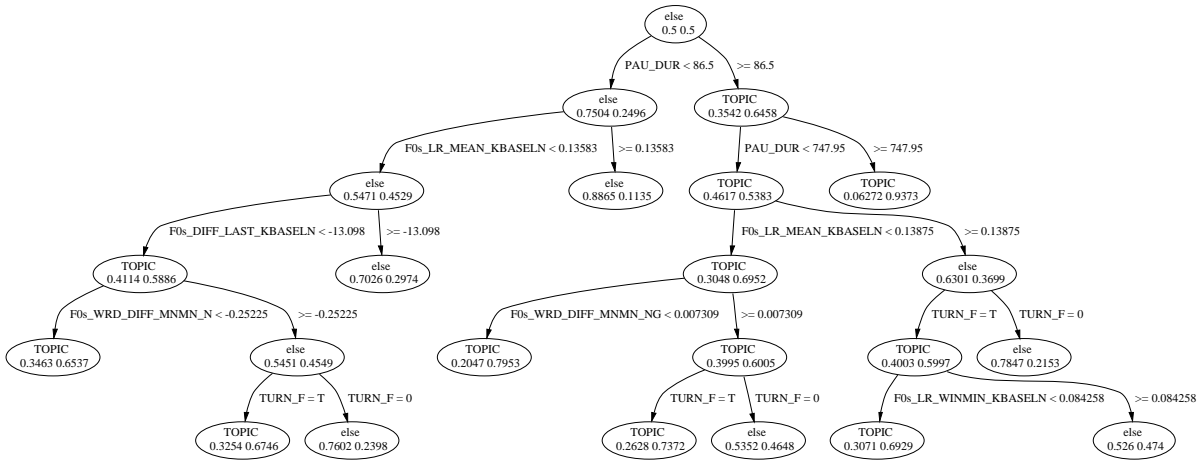,width=\textwidth,angle=-90}
\end{center}
\caption{Top levels of decision tree selected for the Broadcast News topic segmentation task. Nodes contain the percentage of  ``else'' and ``TOPIC''  boundaries, respectively, and are labeled with the majority class.} 
\label{fig:tree-topic-bn} 
\end{figure*}

Inspecting the tree in Fig.~\ref{fig:tree-topic-bn} (the tree has
additional leaves; we show only the top of it), we find that it is
easily interpretable and consistent with prosodic descriptions of
topic or paragraph boundaries. Boundaries are indicated by longer
pauses and by turn information, as expected. Note that the pause
thresholds are considerably higher than those used for the sentence
tree. This is as expected, because of the larger units used here, and
due to the prior chopping at long pause boundaries for this task.

Most of the rest of the tree uses F0 information, in two ways. The
most useful F0 range feature, {\it F0s\_LR\_MEAN\_KBASELN}, computes
the log of the ratio of the mean F0 in the last word to the speaker's
estimated F0 baseline.  As shown, lower values favor topic boundaries,
which is consistent with speakers dropping to the bottom of their
pitch ranges at the ends of topic units. The other F0 feature reflects
the height of the last word relative to a speaker's estimated F0
range; smaller values thus indicate that a speaker is closer to his or her
F0 floor, and as would be predicted, imply topic boundaries.

\begin{figure}
\begin{center}
\strut\psfig{figure=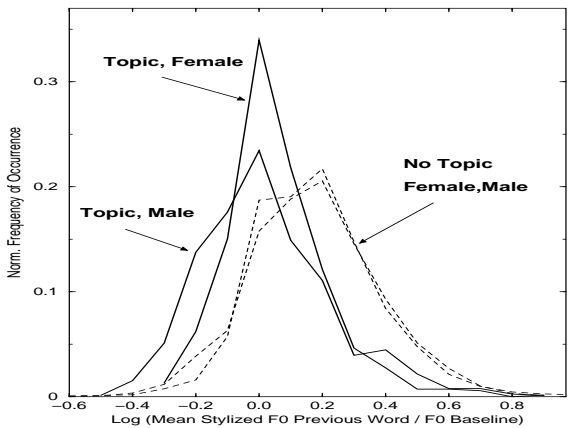,width=\columnwidth,height=0.75\columnwidth}
\end{center}
\caption{Normalized distribution of F0 range feature ({\it F0s\_LR\_MEAN\_KBASELN}) for male and female speakers for topic and nontopic boundaries in Broadcast News}
\label{fig:gender}
\end{figure}

The speaker-gender feature was used in the tree in a pattern that at
first suggested to us a potential problem with our normalizations.  It
was repeatedly used immediately after conditioning on the F0 range
feature {\it F0s\_LR\_MEAN\_KBASELN}.
However, inspection of the feature value distributions by gender and
by boundary class suggested that this was not a problem with normalization,
as shown in Fig.~\ref{fig:gender}.

As indicated, there was no difference by gender in the distribution of
F0 values for the feature in the case of boundaries not containing a
topic change.  After normalization, both men and women ended nontopic
boundaries in similar regions above their baselines.  Since nontopic
boundaries are by far the more frequent class (distributions in the
histogram are normalized), the majority of boundaries in the data show
no difference on this measure by gender.  For topic boundaries,
however, the women in a sense behave more ``neatly'' than the men.  As
a group, the women have a tighter distribution, ending topics at F0
values that are centered closely around their F0 baselines. Men, on the
other hand, are as a group somewhat less ``well-behaved'' in this
regard. They often end topics below their F0 baselines, and showing a
wider distribution (although it should also be noted that since these
are aggregate distributions, the wider distribution for men could
reflect either within-speaker or cross-speaker variation).

This difference is unlikely to be due to baseline estimation problems,
since the nontopic distributions show no difference. The variance
difference is also not explained by a difference in sample size, since
that factor would predict an effect in the opposite direction.  One
possible explanation is that men are more likely than women to produce
regions of nonmodal voicing (such as creak) at the ends of topic
boundaries; this awaits further study.  In addition, we noted that
nontopic pauses (i.e., chopping boundaries) are much more likely to
occur in male than in female speech, a phenomenon that could have
several causes.  For example, it could be that male speakers in
Broadcast News are assigned longer topic segments on average, or that
male speakers are more prone to pausing in general, or that males
dominate the spontaneous speech portions where pausing is naturally
more frequent. This finding, too, awaits further analysis.

\subsubsection{Error reduction from prosody}

Table~\ref{tab:bn-topic-results} shows results for segmentation into
topics in Broadcast News speech.  All results reflect the word-averaged,
weighted error metric used in the TDT-2 evaluations \cite{TDT2}.
Chance here corresponds to outputting the ``no boundary'' class at
all locations, meaning that the false alarm rate will be zero, and the miss
rate will be 1. Since the TDT metric assigns a weight of 0.7 to false alarms,
and 0.3 to misses, chance in this case will be 0.3. 

\begin{table}
\caption{\label{tab:bn-topic-results}
        Results for topic segmentation on Broadcast News}
\begin{center}
\begin{tabular}{lll}
\hline
Model           & Transcribed 		& Recognized \\
		& words			& words \\
\hline
LM only         &       0.1895          & 0.1897        \\
Prosody only    &       0.1657          & 0.1731        \\
Combined    	&       0.1377          & 0.1438        \\
HMM		& 			& \\
		& 			& \\
Chance          &       0.3             & 0.3           \\
\hline
\end{tabular}
\\ Values indicate the TDT weighted segmentation cost metric.
\end{center}
\end{table}

As shown, the error rate for the prosody model alone is lower than
that for the language model. Furthermore, combining the models yields
a significant improvement. Using the combined model, the error rate
decreased by 27.3\% relative to the language model, for the correct
words, and by 24.2\% for recognized words.

\subsubsection{Performance without F0 features} %NEW

As in the earlier case of Broadcast News sentence segmentation, since
this task made use of F0 features, we asked how well it would fare
without any F0 features. The experiments were conducted only for true
words, since as shown previously in Table~\ref{tab:bn-topic-results},
results are similar to those for recognized words.  Results, as shown
in Table~\ref{tab:bn-topic-results-nof0}, indicate a significant
degradation in performance when the F0 features are removed.

\begin{table}
\caption{\label{tab:bn-topic-results-nof0}
        Results for topic segmentation on Broadcast News}
\begin{center}
\begin{tabular}{lll}
\hline
Model           & Transcribed words  \\
\hline
LM only         &       0.1895       \\
		&		\\
%Prosody only    &       0.1657       \\
Combined HMM: & \\
~~~All prosodic features    &       0.1377       \\
~~~No F0 features           &       0.1511       \\
		& 		& \\
Chance          &       0.3        \\
\hline
\end{tabular}
\\ Values indicate the TDT weighted segmentation cost metric.
\end{center}
\end{table}

\subsection{Comparisons of error reduction across conditions}

To compare performance of the prosodic, language, and combined models
directly across tasks and corpora, it is necessary to normalize over
three sources of variation. First, our conditions differ in chance
performance (since the percentage of boundaries that correspond to a
sentence or topic change differ across tasks and copora).  Second, the
upper bound on accuracy in the case of imperfect word recognition
depends on both the word error rate of the recognizer for the corpus,
and the task.  Third, the (standard) metric we have used to evaluate
topic boundary detection differs from the straight accuracy metric
used to assess sentence boundary detection.

A meaningful metric for comparing results directly across tasks is the
percentage of the chance error that remains after application of the
modeling. This measure takes into account the different chance values,
as well as the ceiling effect on accuracy due to recognition errors.
Thus, a model with a score of 1.0 does no better than chance for that
task, since 100\% of the error associated with chance performance
remains after the modeling. A model with a score close to 0.0 is a
nearly ``perfect'' model, since it eliminates nearly all the chance
error.  Note that in the case of recognized words, this amounts to an
error rate at the lower bound rather than at zero.

\begin{figure*}
\begin{center}
\strut\psfig{figure=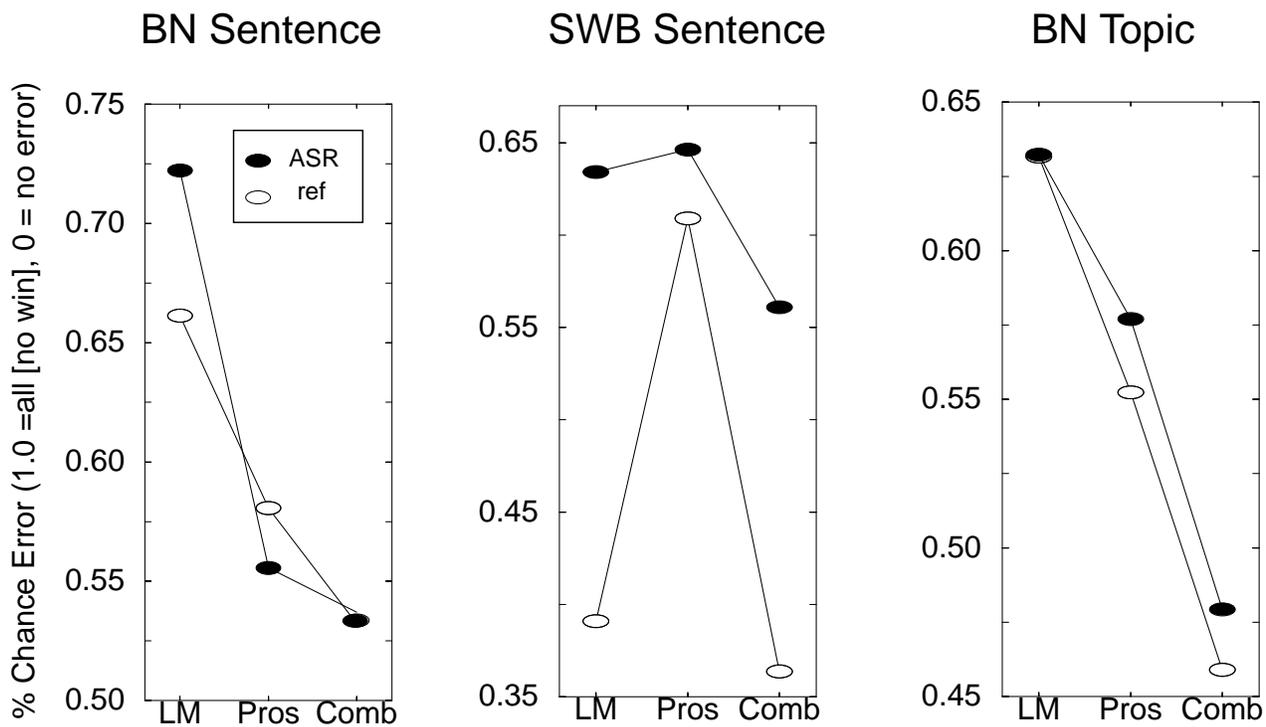,width=4in,angle=270}
\vspace{2.5in}
\end{center}
\caption{Percentage of chance error remaining after application of model (allows performance to be directly compared across tasks).  BN=Broadcast News, SWB=Switchboard, ASR=1-best recognition hypothesis, ref=transcribed words, LM=language model only, Pros=prosody model only, Comb=combination of language and prosody models.} 
\label{fig:rel-error}
\end{figure*}

In Fig.~\ref{fig:rel-error}, performance on the relative error
metric is plotted by task/corpus, reliability of word cues (ASR or
reference transcript), and model. In the case of the combined model,
the plotted value reflects performance for whichever of the two
combination approaches (HMM or interpolation) yielded best results for
that condition.

Useful cross-condition comparisons can be summarized.  For all tasks
and as expected, performance suffers for recognized words compared
with transcribed words.  For the sentence segmentation tasks, the
prosodic model degrades less on recognized words relative to true
words than the word-based models.  The topic segmentation results
based on language model information show remarkable robustness to
recognition errors---much more so than sentence segmentation. This can
be noted by comparing the large loss in performance from reference to
ASR word cues for the language model in the two sentence tasks, to the
identical performance of reference and ASR words in the case of the
topic task.  The pattern of results can be attributed to the different
character of the language model used.  Sentence segmentation uses a
higher-order N-gram that is sensitive to specific words around a
potential boundary, whereas topic segmentation is based on
bag-of-words models that are inherently robust to individual word
errors.

Another important finding made visible in Fig.~\ref{fig:rel-error}
is that the performance of the language model alone on Switchboard
transcriptions is unusually good, when compared with the performance
of the language model alone for all other conditions (including the
corresponding condition for Broadcast News).  This advantage for
Switchboard completely disappears on recognized words.  While
researchers typically have found Switchboard  a difficult corpus
to process, in the case of sentence segmentation on true words it is
just the opposite---atypically easy.  Thus, previous work on automatic
segmentation on Switchboard transcripts \cite{StoShr:icslp96} is
likely to overestimate success for other corpora.  The Switchboard
sentence segmentation advantage is due in large part to the high rate
of a small number of words that occur sentence-initially (especially
``I'', discourse markers, backchannels, coordinating conjunctions, and
disfluencies).

Finally, a potentially interesting pattern can be seen
when comparing the two alternative model combination approaches
(integrated HMM, or interpolation) for the sentence segmentation task.%
\footnote{The interpolated model combination is not possible for topic
  segmentation, as explained earlier.}  Only the best-performing model
combination approach for each condition (ASR or reference words) is
noted in Fig.~\ref{fig:rel-error}; however, the complete set of
results is inferrable from Tables~\ref{tab:bn-sent-results}
and~\ref{tab:swb-sent-results}.  As indicated in the tables, the same
general pattern obtained for both corpora.  The integrated HMM was the
better approach on true words, but it fared relatively poorly on
recognized words.
The posterior interpolation, on the other hand, yielded smaller, but
consistent improvements over the individual knowledge sources on both
true and recognized words.  The pattern deserves further study, but
one possible explanation is that the integrated HMM approach as we
have implemented it assumes that the prosodic features are independent
of the words.  Recognition errors, however, will tend to affect both
words (by definition) and prosodic features through incorrect
alignments.  This will cause the two types of observations to be
correlated, violating the independence assumption.

\subsection{General discussion and future work}

There are a number of ways in which the studies just described could
be improved and extended in future work.  One issue for the prosodic
modeling is that currently, all of our features come from a small
window around the potential boundary.  It is possible that prosodic
properties spanning a longer range could convey additional useful
information.  A second likely source of improvement would be to
utilize information about lexical stress and syllable structure in
defining features (for example, to better predict the domain of
prefinal lengthening).  Third, additional features should be
investigated; in particular it would be worthwhile to examine
energy-related features if effective normalization of channel and
speaker characteristics could be achieved. Fourth, our decision tree
models might be improved by using alternative algorithms to induce
combinations of our basic input features. This could result in smaller
and/or better-performing trees. Finally, as mentioned earlier, testing
on recognized words involved a fundamental mismatch with respect to
model training, where only true words were used. This mismatch worked
against us, since the (fair) testing on recognized words used prosodic
models that had been optimized for alignments from true words.  Full
retraining of all model components on recognized words would be an
ideal (albeit presently expensive) solution to this problem.

Comparisons between the two speech styles in terms of prosodic feature
usage would benefit from a study in which factors such as speaker
overlap in train and test data, and the sound quality of recordings,
are more closely controlled across corpora.  As noted earlier,
Broadcast News had an advantage over Switchboard in terms of speaker
consistency, since as is typical in speech recognition evaluations on
news speech, it included speaker overlap in training and testing. This
factor may have contributed to more robust performance for features
dependent on good speaker normalization---particularly for the F0
features, which used an estimate of the speaker's baseline pitch.  It
is also not yet clear to what extent performance for certain features
is affected by factors such as recording quality and bandwidth, versus
aspects of the speaking style itself. For example, it is possible that
a high-quality, full-bandwidth recording of Switchboard-style speech
would show a greater use of prosodic features than found here.

An added area for further study is to adapt prosodic or language models
to the local context.  For example, Broadcast News exhibits an
interesting variety of shows, speakers, speaking styles, and acoustic
conditions.  Our current models contain only very minimal conditioning
on these local properties.  However, we have found in other work that
tuning the topic segmenter to the type of broadcast show provided
significant improvement \cite{TurEtAl:CL2000}.  The
sentence segmentation task could also benefit from explicit modeling
of speaking style.  For example, our results show that both lexical
and prosodic sentence segmentation cues differ substantially between
spontaneous and planned speech. Finally, results might be improved by
taking advantage of speaker-specific information (i.e. behaviors or
tendencies beyond those accounted for by the speaker-specific
normalizations included in the prosodic modeling).  Initial
experiments suggest we did not have enough training data per speaker
available for an investigation of speaker-specific modeling; however,
this could be made possible through additional data or the use of
smoothing approaches to adapt global models to speaker-specific ones.

More sophisticated model combination approaches that explicitly model
interactions of lexical and prosodic features offer much promise for
future improvements.  Two candidate approaches are the decision trees
based on unsupervised hierarchical word clustering of
\cite{HeemanAllen:97}, and the feature selection approach for
exponential models \cite{Beeferman:99}. As shown in
\namecite{StoShr:icslp96} and similar to \namecite{HeemanAllen:97}, it is
likely that the performance of our segmentation language models would
be improved by moving to an approach based on word classes.

Finally, the approach developed here could be extended to other
languages, as well as to other tasks.  As noted in
Section~\ref{sec:why-use-prosody}, prosody is used across languages to
convey information units (e.g., \cite{Vaissiere:83}, among others).
While there is broad variation across languages in the manner in which
information related to item salience (accentuation and prominence) is
conveyed, there are similarities in many of the features used to
convey boundaries.  Such universals include pausing, pitch declination
(gradual lowering of F0 valleys throughout both sentences and
paragraphs), and amplitude and F0 resets at the beginnings of major
units.  One could thus potentially extend this approach to a new
language.  The prosodic features would differ, but it is expected that
for many languages, similar basic raw features of pausing, duration,
and pitch can be effective in segmentation tasks.  In a similar vein,
although prosodic features depend on the type of events one is trying
to detect, the general approach could be extended to tasks beyond
sentence and topic segmentation (see, for example,
\opencite{DilekEtAl:euro99,ShribergEtAl:LS}).

\section{Summary and conclusion} \label{sec:summary-conclusion}

We have studied the use of prosodic information for sentence and topic
segmentation, both of which are important tasks for information
extraction and archival applications.  Prosodic features reflecting
pause durations, suprasegmental durations, and pitch contours were
automatically extracted, regularized, and normalized. They required no
hand-labeling of prosody; rather, they were based solely on time
alignment information (either from a forced alignment or from
recognition hypotheses).

The features were used as inputs to a decision tree model, which  predicted
the appropriate segment boundary type at each inter-word boundary.  We
compared the performance of these prosodic predictors to that of
statistical language models capturing lexical correlates of segment
boundaries, as well as to combined models integrating both lexical and
prosodic information.  Two knowledge source integration approaches
were investigated: one based on interpolating posterior probability
estimators, and the other using a combined HMM that emitted both
lexical and prosodic observations.

Results showed that on Broadcast News the prosodic model alone
performed as well as (or even better than) purely word-based
statistical language models, for both true and automatically
recognized words.  The prosodic model achieved comparable performance
with significantly less training data, and often degraded less due to
recognition errors.  Furthermore, for all tasks and corpora, we obtained
a significant improvement over word-only models using one or both of our
combined models.  Interestingly, the integrated HMM worked best on
transcribed words, while the posterior interpolation approach was much
more robust in the case of recognized words.

Analysis of the prosodic decision trees revealed that the models
capture language-independent boundary indicators described in the
literature, such as preboundary lengthening, boundary tones, and pitch
resets.  Consistent with descriptive work, larger breaks such as
topics, showed features similar to those of sentence breaks, but with
more pronounced pause and intonation patterns.  Feature usage,
however, was corpus dependent.  While features such as pauses were
heavily used in both corpora, we found that pitch is a highly
informative feature in Broadcast News, whereas duration and word cues
dominated in Switchboard.  We conclude that prosody provides rich and
complementary information to lexical information for the detection of
sentence and topic boundaries in different speech styles, and that it
can therefore play an important role in the automatic segmentation of
spoken language.

\section*{Acknowledgements}

We thank Kemal S\"onmez for providing the model for F0 stylization
used in this work; Rebecca Bates, Mari Ostendorf, Ze'ev Rivlin, Ananth
Sankar, and Kemal S\"onmez for invaluable assistance in data
preparation and discussions; Madelaine Plauch{\'e} for hand-checking
of F0 stylization output and regions of nonmodal voicing; and Klaus
Ries, Paul Taylor, and an anonymous reviewer for helpful comments on
earlier drafts.  This research was supported by DARPA under contract
no.~N66001-97-C-8544 and by NSF under STIMULATE grant IRI-9619921. The
views herein are those of the authors and should not be interpreted as
representing the policies of the funding agencies.

\footnotesize

\bibliographystyle{myapa}
% \bibliography{/home/speech/stolcke/lib/bib/all,prosody}
\bibliography{all,prosody}

\end{document}